\documentclass[10pt]{article} % For LaTeX2e
\usepackage[preprint]{tmlr}

\usepackage{hyperref}
\usepackage{url}
\usepackage{amssymb}
\usepackage{amsfonts}
\usepackage{amsmath}
\usepackage{mathtools}
\usepackage{mathrsfs}
\usepackage{graphicx}
\usepackage{subfig}
\usepackage{algorithm}
\usepackage{algpseudocode}
\usepackage{blindtext}
\usepackage{titlesec}
\usepackage[nottoc,numbib]{tocbibind}

\usepackage{amsmath,amsfonts,bm}

\def\eqref#1{equation~\ref{#1}}
\def\1{\bm{1}}

\DeclareMathAlphabet{\mathsfit}{\encodingdefault}{\sfdefault}{m}{sl}
\SetMathAlphabet{\mathsfit}{bold}{\encodingdefault}{\sfdefault}{bx}{n}

\usepackage{hyperref}
\usepackage{url}

\title{A Technical Note on the Architectural Effects on Maximum Dependency Lengths of Recurrent Neural Networks}

\author{\name Jonathan S. Kent \email jonathan.s.kent@lmco.com \\
      \addr Advanced Technology Center\\
      Lockheed Martin Space;\\
      Columbia University
      \AND
      \name Michael M. Murray \email michael.m.murray@lmco.com \\
      \addr Advanced Technology Center \\
      Lockheed Martin Space}

\def\month{MM}  % Insert correct month for camera-ready version
\def\year{YYYY} % Insert correct year for camera-ready version
\def\openreview{\url{https://openreview.net/forum?id=XXXX}} % Insert correct link to OpenReview for camera-ready version

\begin{document}

\maketitle

\begin{abstract}
This work proposes a methodology for determining the maximum dependency length of a recurrent neural network (RNN), and then studies the effects of architectural changes, including the number and neuron count of layers, on the maximum dependency lengths of traditional RNN, gated recurrent unit (GRU), and long-short term memory (LSTM) models.
\end{abstract}

\section{Introduction}

Recurrent neural networks (RNNs) are a class of artificial neural network that, by maintaining and modifying an internal state over sequential inputs, are able to model dependencies over the course of the sequence \cite{medsker1999recurrent, salehinejad2017recent}. This makes them useful for signal processing tasks. Gated recurrent units (GRUs) are a subtype of RNNs, which introduce a gating mechanism to try and improve the models' abilities to maintain a coherent and information-rich hidden state over time \cite{cho2014learning}. Long-short term memory models (LSTMs) are another subtype, which expand on the complexity of the hidden state mechanisms in order to achieve longer maximum dependencies, in part by reducing the effects of vanishing gradients \cite{10.5555/2998981.2999048}.

Although any RNN model can express arbitrarily long dependencies theoretically or via manual construction, practically speaking the ability for models to learn these dependencies from data varies greatly depending on the model architecture. In order to effectively use these models for appropriate application, with enough capacity to be useful but not so much as to be wasteful, it can be helpful for practitioners to have heuristic estimates for the effects of changing the number of neurons and the number of layers in an RNN model, or, inversely, how many layers and neurons will be needed to achieve a particular result. We now introduce a method for measuring the maximum dependency length of an RNN given its architecture, using domain-agnostic synthetic data.

\section{Methodology}

Our method for determining the maximum dependency length of a neural network involves the delayed reconstruction of a square wave. This task is simple enough that it should measure only the dependency length, and not factors related to the complexity of the learned function, like replicating complex patterns or modeling language.

We begin by generating input binary square waves with a value of 0 for all time steps that are not part of the wave, and 1 for a contiguous subsequence $d = 5$ time steps in length. A second ground truth square wave is generated, identical to the first, except with a delay of $l$ time steps between the last 1-valued time step of the input wave and the first 1-valued time step of the output wave. For a value of $l = 3$, examples are provided in Table \ref{tab:wex}.

\begin{table}[!h]
    \centering
    \begin{tabular}{c|l}
      Input  &\texttt{00000011111000000000000}\\
      Ground Truth  &\texttt{00000000000000111110000}
    \end{tabular}
    \caption{Example Waveforms}
    \label{tab:wex}
\end{table}

Batches of these input/output sequence pairs are generated with a random delay from the first time step to the start of the input wave, and provided to each RNN model for training. Each RNN model includes a fully connected output module, which takes the output of the RNN at each time step, and reduces the dimensionality of the output to 1. The sign of the output is then taken, and interpreted as 0 where negative, and 1 where positive. Cross entropy loss is used for training, as well as the Adam optimizer \cite{kingma2017adam}. A successful wave replication is recorded when 90\% of the ground truth 1-values are matched, as are more than 95\% of the ground truth 0-values. Further training details are provided in Appendix \ref{app:trdet}. When every wave in a batch is recorded as successful, the model is considered to have succeeded once for that value of $l$, being the dependency length.

For a given value of $l$, models are trained from scratch five times, with at least four eventual successes resulting in the over-all assessment that the model is capable of learning that dependency length. Each model starts with an attempt to learn with $l = 1$; a positive assessment results in the value of $l$ doubling, and a negative assessment resulting in the commencement of a binary search between that value and one half of that value of $l$, for the threshold such that the model is assessed positively at that value, but assessed negatively for all values greater. This grow-terminate-search procedure is run five times for each model architecture, and the results recorded.

We experiment with PyTorch default implementations of RNNs, GRUs, and LSTMs, with 8 to 128 neurons per layer in increments of 8, and 1 to 8 layers, for a total of 384 models, with 1,920 total executions of the grow-terminate-search procedure. All numerical results are provided in Appendix \ref{app:expr}, including the minimum, median, mean, and maximum values, and standard deviations, in their own columns.

\section{Results}

\begin{figure}[!h]
  \centering
  \subfloat{\includegraphics[width=0.32\textwidth]{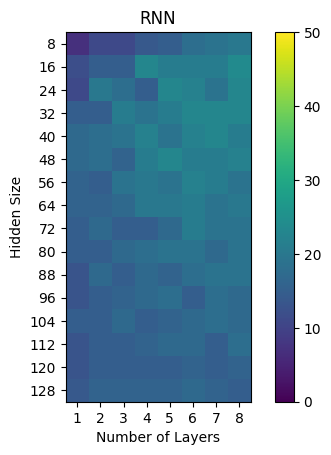}}
  \hfill
  \subfloat{\includegraphics[width=0.32\textwidth]{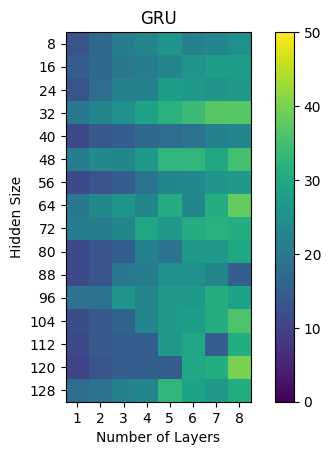}}
  \hfill
  \subfloat{\includegraphics[width=0.32\textwidth]{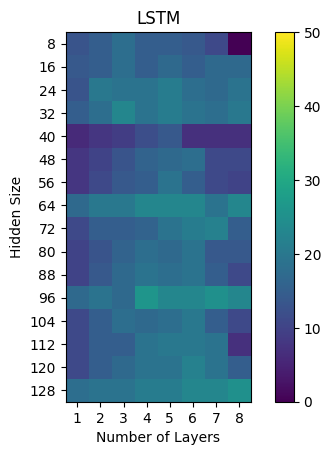}}
  \caption{Graphs of the minimum value over five runs for each model.}
  \label{fig:mingraphs}
\end{figure}

\begin{figure}[!h]
  \centering
  \subfloat{\includegraphics[width=0.32\textwidth]{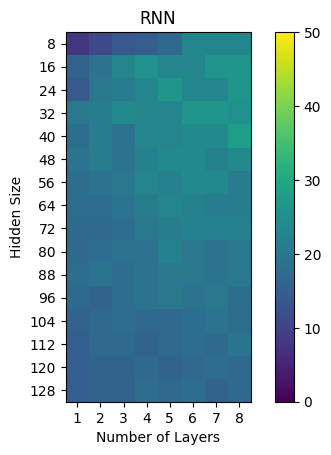}}
  \hfill
  \subfloat{\includegraphics[width=0.32\textwidth]{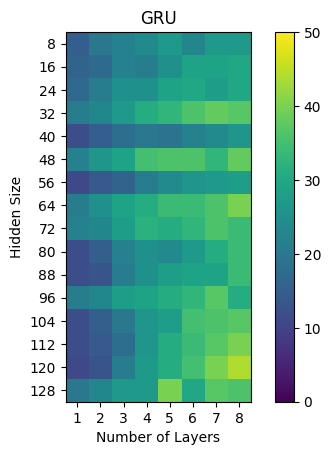}}
  \hfill
  \subfloat{\includegraphics[width=0.32\textwidth]{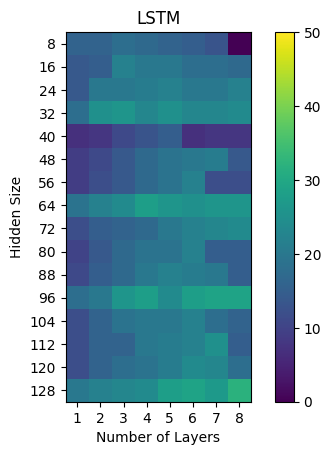}}
  \caption{Graphs of the median value over five runs for each model.}
  \label{fig:medgraphs}
\end{figure}

\begin{figure}[!h]
  \centering
  \subfloat{\includegraphics[width=0.32\textwidth]{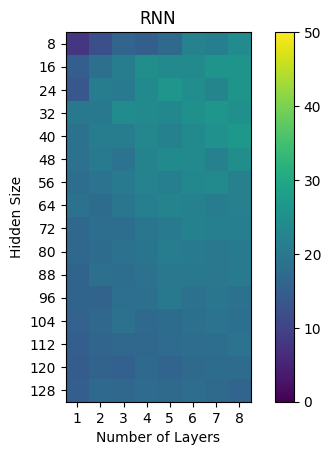}}
  \hfill
  \subfloat{\includegraphics[width=0.32\textwidth]{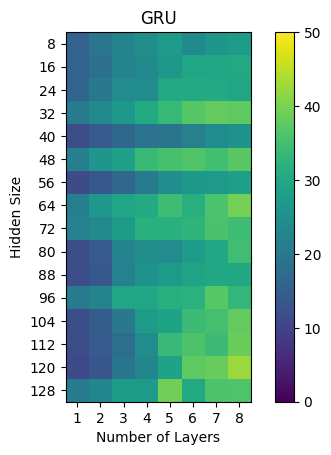}}
  \hfill
  \subfloat{\includegraphics[width=0.32\textwidth]{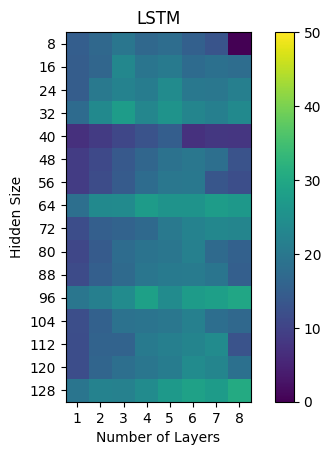}}
  \caption{Graphs of the mean value over five runs for each model.}
  \label{fig:meangraphs}
\end{figure}

\begin{figure}[!h]
  \centering
  \subfloat{\includegraphics[width=0.32\textwidth]{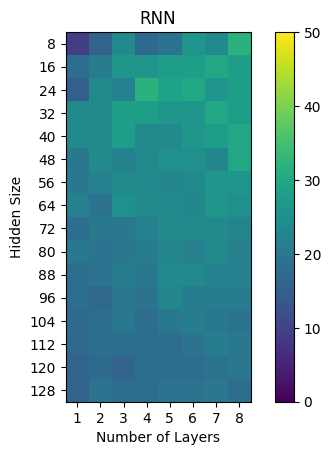}}
  \hfill
  \subfloat{\includegraphics[width=0.32\textwidth]{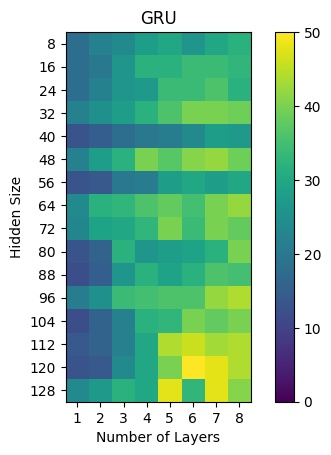}}
  \hfill
  \subfloat{\includegraphics[width=0.32\textwidth]{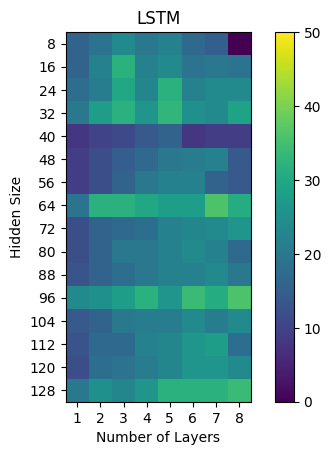}}
  \caption{Graphs of the maximum value over five runs for each model.}
  \label{fig:maxgraphs}
\end{figure}

We now present the results of the experiments, graphed in Figures \ref{fig:mingraphs}, \ref{fig:medgraphs}, \ref{fig:meangraphs}, \ref{fig:maxgraphs}. Although the results may seem counter-intuitive, we would like to make a note that these results interrelate maximum dependency length with the training requirements for each model, given that all models were trained using the same procedure and hyperparameters. Thus, a model may have greater training requirements to reach its full potential, compared to that with which it was provided. The maximum achieved value was a GRU with 6 layers, 120 neurons each, with a dependency length of 50.

These results include a large amount of statistical noise, as well as artifacts assumed to relate to the back-end implementations. These are left in, both to show the limitations of the approach, and to highlight what results may look like for practitioners approaching related problems; there will always be some amount of noise in the training pipeline, and there will always be artifacts of the implementation used. This manuscript should not be taken as a purely scientific investigation, but rather as a technical note providing these results.

\clearpage

\clearpage

\clearpage

\bibliography{main}

\begin{thebibliography}{5}
\providecommand{\natexlab}[1]{#1}
\providecommand{\url}[1]{\texttt{#1}}
\expandafter\ifx\csname urlstyle\endcsname\relax
  \providecommand{\doi}[1]{doi: #1}\else
  \providecommand{\doi}{doi: \begingroup \urlstyle{rm}\Url}\fi

\bibitem[Cho et~al.(2014)Cho, Van~Merri{\"e}nboer, Gulcehre, Bahdanau, Bougares, Schwenk, and Bengio]{cho2014learning}
Kyunghyun Cho, Bart Van~Merri{\"e}nboer, Caglar Gulcehre, Dzmitry Bahdanau, Fethi Bougares, Holger Schwenk, and Yoshua Bengio.
\newblock Learning phrase representations using rnn encoder-decoder for statistical machine translation.
\newblock \emph{arXiv preprint arXiv:1406.1078}, 2014.

\bibitem[Hochreiter \& Schmidhuber(1996)Hochreiter and Schmidhuber]{10.5555/2998981.2999048}
Sepp Hochreiter and J\"{u}rgen Schmidhuber.
\newblock Lstm can solve hard long time lag problems.
\newblock In \emph{Proceedings of the 9th International Conference on Neural Information Processing Systems}, NIPS'96, pp.\  473–479, Cambridge, MA, USA, 1996. MIT Press.

\bibitem[Kingma \& Ba(2017)Kingma and Ba]{kingma2017adam}
Diederik~P. Kingma and Jimmy Ba.
\newblock Adam: A method for stochastic optimization, 2017.

\bibitem[Medsker \& Jain(1999)Medsker and Jain]{medsker1999recurrent}
Larry Medsker and Lakhmi~C Jain.
\newblock \emph{Recurrent neural networks: design and applications}.
\newblock CRC press, 1999.

\bibitem[Salehinejad et~al.(2017)Salehinejad, Sankar, Barfett, Colak, and Valaee]{salehinejad2017recent}
Hojjat Salehinejad, Sharan Sankar, Joseph Barfett, Errol Colak, and Shahrokh Valaee.
\newblock Recent advances in recurrent neural networks.
\newblock \emph{arXiv preprint arXiv:1801.01078}, 2017.

\end{thebibliography}
\bibliographystyle{tmlr}

\appendix
\section{Training Details}
\label{app:trdet}

For the purposes of training, toy datasets are created with 1,024 samples, each sample consisting of a pair of input and ground truth waveforms, represented as vectors of dimension $m$, where $m$ is initialized as 100, but may be increased over time to account for large achieved values of $l$ - in practice this does not occur. For each sample, the start time $s$ is randomly selected as an integer from $[0, m - 2d + l]$, where $d = 5$ is the length of the square wave. The input square wave is set to a value of 1 from time $s$ to $s + d$, and the ground truth square wave is set to a value of 1 from time $s + l + d$ to $s + l + 2d$.

Each RNN model is composed of two modules, the RNN itself, and a fully connected network consisting of seven layers, every internal layer of which has 128 neurons, with the last layer having one neuron, and the first layer having as many inputs as the RNN module has neurons. Between every linear layer is a 1D batchnorm, followed by a ReLU activation. At every time step, the RNN model takes the current value of the input waveform as its sole input, updates its hidden state, and provides the hidden state to the fully connected network, which produces its output for that time step.

The loss function is a weighted entropy loss. Probabilities are taken using the sigmoid of the output of the model, which are then compared with the ground truth waveform in the ordinary manner, with the caveat that the loss associated with positive values has a coefficient of $1/d$, and the loss assocated with negative values has a coefficient of $1/(m-d)$.

Each training session consists of a maximum of 200 epochs of Adam optimization, with a learning rate of $10^{-5}$, and a coefficient of weight decay of $10^{-6}$. Training will quit early if success is achieved on a given batch. Samples from the toy dataset are shuffled with each epoch, and provided in batches of size 64.

\clearpage

\section{Numerical Results}
\label{app:expr}

\begin{table}[!b]
    \centering
    \begin{tabular}{c|c|c||c|c|c|c|c||c|c|c|c|c}
        Type & Hidden Size & Layers & R1 & R2 & R3 & R4 & R5 & Min. & Med. & Mean & Max. & Std. \\
        \hline
        RNN & 8 & 1 & 7 & 9 & 9 & 8 & 7 & 7 & 8.0 & 8.0 & 9 & 1.00\\
RNN & 8 & 2 & 11 & 16 & 11 & 11 & 12 & 11 & 11.0 & 12.2 & 16 & 2.17\\
RNN & 8 & 3 & 13 & 20 & 14 & 24 & 11 & 11 & 14.0 & 16.4 & 24 & 5.41\\
RNN & 8 & 4 & 14 & 15 & 17 & 15 & 15 & 14 & 15.0 & 15.2 & 17 & 1.10\\
RNN & 8 & 5 & 15 & 17 & 17 & 19 & 18 & 15 & 17.0 & 17.2 & 19 & 1.48\\
RNN & 8 & 6 & 23 & 23 & 26 & 18 & 21 & 18 & 23.0 & 22.2 & 26 & 2.95\\
RNN & 8 & 7 & 23 & 23 & 19 & 19 & 24 & 19 & 23.0 & 21.6 & 24 & 2.41\\
RNN & 8 & 8 & 21 & 20 & 25 & 32 & 23 & 20 & 23.0 & 24.2 & 32 & 4.76\\
\hline
RNN & 16 & 1 & 18 & 16 & 12 & 13 & 16 & 12 & 16.0 & 15.0 & 18 & 2.45\\
RNN & 16 & 2 & 19 & 20 & 21 & 15 & 17 & 15 & 19.0 & 18.4 & 21 & 2.41\\
RNN & 16 & 3 & 15 & 24 & 26 & 18 & 23 & 15 & 23.0 & 21.2 & 26 & 4.55\\
RNN & 16 & 4 & 23 & 25 & 26 & 26 & 24 & 23 & 25.0 & 24.8 & 26 & 1.30\\
RNN & 16 & 5 & 21 & 21 & 23 & 26 & 28 & 21 & 23.0 & 23.8 & 28 & 3.11\\
RNN & 16 & 6 & 26 & 23 & 21 & 22 & 28 & 21 & 23.0 & 24.0 & 28 & 2.92\\
RNN & 16 & 7 & 21 & 26 & 24 & 28 & 30 & 21 & 26.0 & 25.8 & 30 & 3.49\\
RNN & 16 & 8 & 24 & 28 & 25 & 26 & 26 & 24 & 26.0 & 25.8 & 28 & 1.48\\
\hline
RNN & 24 & 1 & 13 & 15 & 15 & 14 & 11 & 11 & 14.0 & 13.6 & 15 & 1.67\\
RNN & 24 & 2 & 20 & 24 & 22 & 20 & 20 & 20 & 20.0 & 21.2 & 24 & 1.79\\
RNN & 24 & 3 & 21 & 18 & 22 & 22 & 21 & 18 & 21.0 & 20.8 & 22 & 1.64\\
RNN & 24 & 4 & 15 & 22 & 32 & 23 & 28 & 15 & 23.0 & 24.0 & 32 & 6.44\\
RNN & 24 & 5 & 29 & 23 & 26 & 26 & 26 & 23 & 26.0 & 26.0 & 29 & 2.12\\
RNN & 24 & 6 & 25 & 23 & 23 & 30 & 22 & 22 & 23.0 & 24.6 & 30 & 3.21\\
RNN & 24 & 7 & 22 & 23 & 26 & 24 & 19 & 19 & 23.0 & 22.8 & 26 & 2.59\\
RNN & 24 & 8 & 25 & 26 & 23 & 27 & 28 & 23 & 26.0 & 25.8 & 28 & 1.92\\
\hline
RNN & 32 & 1 & 24 & 15 & 18 & 24 & 20 & 15 & 20.0 & 20.2 & 24 & 3.90\\
RNN & 32 & 2 & 21 & 24 & 24 & 15 & 17 & 15 & 21.0 & 20.2 & 24 & 4.09\\
RNN & 32 & 3 & 24 & 21 & 25 & 28 & 23 & 21 & 24.0 & 24.2 & 28 & 2.59\\
RNN & 32 & 4 & 21 & 23 & 28 & 19 & 28 & 19 & 23.0 & 23.8 & 28 & 4.09\\
RNN & 32 & 5 & 26 & 21 & 21 & 23 & 26 & 21 & 23.0 & 23.4 & 26 & 2.51\\
RNN & 32 & 6 & 23 & 26 & 26 & 26 & 24 & 23 & 26.0 & 25.0 & 26 & 1.41\\
RNN & 32 & 7 & 27 & 26 & 25 & 30 & 23 & 23 & 26.0 & 26.2 & 30 & 2.59\\
RNN & 32 & 8 & 23 & 23 & 25 & 28 & 26 & 23 & 25.0 & 25.0 & 28 & 2.12\\
\hline
RNN & 40 & 1 & 24 & 18 & 17 & 17 & 18 & 17 & 18.0 & 18.8 & 24 & 2.95\\
RNN & 40 & 2 & 19 & 21 & 24 & 18 & 24 & 18 & 21.0 & 21.2 & 24 & 2.77\\
RNN & 40 & 3 & 20 & 19 & 28 & 19 & 19 & 19 & 19.0 & 21.0 & 28 & 3.94\\
RNN & 40 & 4 & 22 & 24 & 23 & 24 & 23 & 22 & 23.0 & 23.2 & 24 & 0.84\\
RNN & 40 & 5 & 24 & 23 & 19 & 21 & 23 & 19 & 23.0 & 22.0 & 24 & 2.00\\
RNN & 40 & 6 & 22 & 25 & 24 & 26 & 23 & 22 & 24.0 & 24.0 & 26 & 1.58\\
RNN & 40 & 7 & 24 & 23 & 28 & 27 & 24 & 23 & 24.0 & 25.2 & 28 & 2.17\\
RNN & 40 & 8 & 26 & 21 & 30 & 28 & 28 & 21 & 28.0 & 26.6 & 30 & 3.44\\
\hline
RNN & 48 & 1 & 19 & 17 & 18 & 20 & 20 & 17 & 19.0 & 18.8 & 20 & 1.30\\
RNN & 48 & 2 & 21 & 22 & 18 & 18 & 24 & 18 & 21.0 & 20.6 & 24 & 2.61\\
RNN & 48 & 3 & 22 & 19 & 16 & 19 & 19 & 16 & 19.0 & 19.0 & 22 & 2.12\\
RNN & 48 & 4 & 24 & 21 & 23 & 22 & 22 & 21 & 22.0 & 22.4 & 24 & 1.14\\
RNN & 48 & 5 & 24 & 24 & 24 & 25 & 23 & 23 & 24.0 & 24.0 & 25 & 0.71\\
RNN & 48 & 6 & 24 & 25 & 25 & 24 & 21 & 21 & 24.0 & 23.8 & 25 & 1.64\\
RNN & 48 & 7 & 21 & 21 & 23 & 23 & 22 & 21 & 22.0 & 22.0 & 23 & 1.00\\
RNN & 48 & 8 & 24 & 24 & 22 & 30 & 24 & 22 & 24.0 & 24.8 & 30 & 3.03\\
    \end{tabular}
\end{table}

\clearpage

\begin{table}[]
    \centering
    \begin{tabular}{c|c|c||c|c|c|c|c||c|c|c|c|c}
        Type & Hidden Size & Layers & R1 & R2 & R3 & R4 & R5 & Min. & Med. & Mean & Max. & Std. \\
        \hline
        RNN & 56 & 1 & 20 & 17 & 18 & 16 & 19 & 16 & 18.0 & 18.0 & 20 & 1.58\\
RNN & 56 & 2 & 18 & 22 & 21 & 15 & 19 & 15 & 19.0 & 19.0 & 22 & 2.74\\
RNN & 56 & 3 & 24 & 19 & 20 & 21 & 20 & 19 & 20.0 & 20.8 & 24 & 1.92\\
RNN & 56 & 4 & 24 & 23 & 20 & 23 & 21 & 20 & 23.0 & 22.2 & 24 & 1.64\\
RNN & 56 & 5 & 21 & 23 & 19 & 22 & 23 & 19 & 22.0 & 21.6 & 23 & 1.67\\
RNN & 56 & 6 & 22 & 24 & 24 & 24 & 23 & 22 & 24.0 & 23.4 & 24 & 0.89\\
RNN & 56 & 7 & 21 & 26 & 26 & 24 & 23 & 21 & 24.0 & 24.0 & 26 & 2.12\\
RNN & 56 & 8 & 21 & 26 & 19 & 20 & 23 & 19 & 21.0 & 21.8 & 26 & 2.77\\
\hline
RNN & 64 & 1 & 17 & 22 & 16 & 18 & 20 & 16 & 18.0 & 18.6 & 22 & 2.41\\
RNN & 64 & 2 & 19 & 17 & 16 & 18 & 18 & 16 & 18.0 & 17.6 & 19 & 1.14\\
RNN & 64 & 3 & 18 & 25 & 19 & 17 & 19 & 17 & 19.0 & 19.6 & 25 & 3.13\\
RNN & 64 & 4 & 22 & 24 & 21 & 21 & 20 & 20 & 21.0 & 21.6 & 24 & 1.52\\
RNN & 64 & 5 & 20 & 24 & 24 & 21 & 23 & 20 & 23.0 & 22.4 & 24 & 1.82\\
RNN & 64 & 6 & 21 & 23 & 22 & 22 & 22 & 21 & 22.0 & 22.0 & 23 & 0.71\\
RNN & 64 & 7 & 21 & 21 & 19 & 26 & 19 & 19 & 21.0 & 21.2 & 26 & 2.86\\
RNN & 64 & 8 & 20 & 23 & 21 & 25 & 20 & 20 & 21.0 & 21.8 & 25 & 2.17\\
\hline
RNN & 72 & 1 & 15 & 17 & 17 & 18 & 17 & 15 & 17.0 & 16.8 & 18 & 1.10\\
RNN & 72 & 2 & 17 & 20 & 19 & 17 & 17 & 17 & 17.0 & 18.0 & 20 & 1.41\\
RNN & 72 & 3 & 18 & 15 & 19 & 20 & 18 & 15 & 18.0 & 18.0 & 20 & 1.87\\
RNN & 72 & 4 & 20 & 22 & 20 & 21 & 15 & 15 & 20.0 & 19.6 & 22 & 2.70\\
RNN & 72 & 5 & 24 & 21 & 17 & 20 & 21 & 17 & 21.0 & 20.6 & 24 & 2.51\\
RNN & 72 & 6 & 22 & 21 & 22 & 24 & 21 & 21 & 22.0 & 22.0 & 24 & 1.22\\
RNN & 72 & 7 & 19 & 19 & 22 & 24 & 24 & 19 & 22.0 & 21.6 & 24 & 2.51\\
RNN & 72 & 8 & 23 & 22 & 21 & 23 & 19 & 19 & 22.0 & 21.6 & 23 & 1.67\\
\hline
RNN & 80 & 1 & 20 & 15 & 17 & 17 & 15 & 15 & 17.0 & 16.8 & 20 & 2.05\\
RNN & 80 & 2 & 19 & 17 & 19 & 15 & 18 & 15 & 18.0 & 17.6 & 19 & 1.67\\
RNN & 80 & 3 & 20 & 19 & 17 & 19 & 18 & 17 & 19.0 & 18.6 & 20 & 1.14\\
RNN & 80 & 4 & 19 & 18 & 19 & 21 & 20 & 18 & 19.0 & 19.4 & 21 & 1.14\\
RNN & 80 & 5 & 22 & 20 & 22 & 23 & 19 & 19 & 22.0 & 21.2 & 23 & 1.64\\
RNN & 80 & 6 & 21 & 20 & 22 & 19 & 20 & 19 & 20.0 & 20.4 & 22 & 1.14\\
RNN & 80 & 7 & 19 & 21 & 17 & 24 & 19 & 17 & 19.0 & 20.0 & 24 & 2.65\\
RNN & 80 & 8 & 22 & 20 & 21 & 20 & 19 & 19 & 20.0 & 20.4 & 22 & 1.14\\
\hline
RNN & 88 & 1 & 13 & 18 & 18 & 18 & 14 & 13 & 18.0 & 16.2 & 18 & 2.49\\
RNN & 88 & 2 & 19 & 18 & 17 & 19 & 19 & 17 & 19.0 & 18.4 & 19 & 0.89\\
RNN & 88 & 3 & 18 & 15 & 19 & 17 & 21 & 15 & 18.0 & 18.0 & 21 & 2.24\\
RNN & 88 & 4 & 20 & 18 & 17 & 19 & 19 & 17 & 19.0 & 18.6 & 20 & 1.14\\
RNN & 88 & 5 & 21 & 19 & 24 & 16 & 20 & 16 & 20.0 & 20.0 & 24 & 2.92\\
RNN & 88 & 6 & 24 & 20 & 20 & 19 & 18 & 18 & 20.0 & 20.2 & 24 & 2.28\\
RNN & 88 & 7 & 20 & 22 & 19 & 19 & 19 & 19 & 19.0 & 19.8 & 22 & 1.30\\
RNN & 88 & 8 & 22 & 20 & 20 & 21 & 19 & 19 & 20.0 & 20.4 & 22 & 1.14\\
\hline
RNN & 96 & 1 & 16 & 18 & 17 & 13 & 17 & 13 & 17.0 & 16.2 & 18 & 1.92\\
RNN & 96 & 2 & 16 & 15 & 16 & 17 & 16 & 15 & 16.0 & 16.0 & 17 & 0.71\\
RNN & 96 & 3 & 16 & 20 & 20 & 17 & 18 & 16 & 18.0 & 18.2 & 20 & 1.79\\
RNN & 96 & 4 & 19 & 17 & 18 & 19 & 19 & 17 & 19.0 & 18.4 & 19 & 0.89\\
RNN & 96 & 5 & 18 & 21 & 19 & 23 & 20 & 18 & 20.0 & 20.2 & 23 & 1.92\\
RNN & 96 & 6 & 19 & 19 & 20 & 21 & 15 & 15 & 19.0 & 18.8 & 21 & 2.28\\
RNN & 96 & 7 & 20 & 19 & 21 & 18 & 20 & 18 & 20.0 & 19.6 & 21 & 1.14\\
RNN & 96 & 8 & 17 & 20 & 17 & 21 & 18 & 17 & 18.0 & 18.6 & 21 & 1.82\\
    \end{tabular}
\end{table}

\clearpage

\begin{table}[]
    \centering
    \begin{tabular}{c|c|c||c|c|c|c|c||c|c|c|c|c}
        Type & Hidden Size & Layers & R1 & R2 & R3 & R4 & R5 & Min. & Med. & Mean & Max. & Std. \\
        \hline
RNN & 104 & 1 & 15 & 16 & 15 & 16 & 17 & 15 & 16.0 & 15.8 & 17 & 0.84\\
RNN & 104 & 2 & 18 & 18 & 15 & 17 & 16 & 15 & 17.0 & 16.8 & 18 & 1.30\\
RNN & 104 & 3 & 20 & 18 & 18 & 17 & 20 & 17 & 18.0 & 18.6 & 20 & 1.34\\
RNN & 104 & 4 & 15 & 18 & 18 & 17 & 17 & 15 & 17.0 & 17.0 & 18 & 1.22\\
RNN & 104 & 5 & 18 & 16 & 17 & 17 & 20 & 16 & 17.0 & 17.6 & 20 & 1.52\\
RNN & 104 & 6 & 18 & 17 & 17 & 21 & 19 & 17 & 18.0 & 18.4 & 21 & 1.67\\
RNN & 104 & 7 & 19 & 18 & 19 & 20 & 19 & 18 & 19.0 & 19.0 & 20 & 0.71\\
RNN & 104 & 8 & 18 & 18 & 17 & 19 & 19 & 17 & 18.0 & 18.2 & 19 & 0.84\\
\hline
RNN & 112 & 1 & 16 & 14 & 13 & 17 & 15 & 13 & 15.0 & 15.0 & 17 & 1.58\\
RNN & 112 & 2 & 15 & 15 & 18 & 17 & 17 & 15 & 17.0 & 16.4 & 18 & 1.34\\
RNN & 112 & 3 & 18 & 17 & 18 & 16 & 15 & 15 & 17.0 & 16.8 & 18 & 1.30\\
RNN & 112 & 4 & 16 & 16 & 16 & 18 & 18 & 16 & 16.0 & 16.8 & 18 & 1.10\\
RNN & 112 & 5 & 17 & 18 & 18 & 17 & 17 & 17 & 17.0 & 17.4 & 18 & 0.55\\
RNN & 112 & 6 & 18 & 17 & 17 & 19 & 18 & 17 & 18.0 & 17.8 & 19 & 0.84\\
RNN & 112 & 7 & 15 & 21 & 17 & 20 & 17 & 15 & 17.0 & 18.0 & 21 & 2.45\\
RNN & 112 & 8 & 18 & 20 & 20 & 18 & 19 & 18 & 19.0 & 19.0 & 20 & 1.00\\
\hline
RNN & 120 & 1 & 16 & 13 & 16 & 15 & 14 & 13 & 15.0 & 14.8 & 16 & 1.30\\
RNN & 120 & 2 & 16 & 17 & 16 & 16 & 15 & 15 & 16.0 & 16.0 & 17 & 0.71\\
RNN & 120 & 3 & 15 & 15 & 16 & 16 & 16 & 15 & 16.0 & 15.6 & 16 & 0.55\\
RNN & 120 & 4 & 17 & 18 & 18 & 17 & 15 & 15 & 17.0 & 17.0 & 18 & 1.22\\
RNN & 120 & 5 & 18 & 15 & 16 & 16 & 17 & 15 & 16.0 & 16.4 & 18 & 1.14\\
RNN & 120 & 6 & 17 & 17 & 18 & 16 & 18 & 16 & 17.0 & 17.2 & 18 & 0.84\\
RNN & 120 & 7 & 15 & 19 & 18 & 18 & 18 & 15 & 18.0 & 17.6 & 19 & 1.52\\
RNN & 120 & 8 & 16 & 17 & 17 & 18 & 20 & 16 & 17.0 & 17.6 & 20 & 1.52\\
\hline
RNN & 128 & 1 & 14 & 15 & 15 & 15 & 16 & 14 & 15.0 & 15.0 & 16 & 0.71\\
RNN & 128 & 2 & 18 & 16 & 19 & 16 & 16 & 16 & 16.0 & 17.0 & 19 & 1.41\\
RNN & 128 & 3 & 18 & 16 & 18 & 16 & 16 & 16 & 16.0 & 16.8 & 18 & 1.10\\
RNN & 128 & 4 & 17 & 18 & 16 & 18 & 18 & 16 & 18.0 & 17.4 & 18 & 0.89\\
RNN & 128 & 5 & 19 & 16 & 18 & 17 & 16 & 16 & 17.0 & 17.2 & 19 & 1.30\\
RNN & 128 & 6 & 17 & 18 & 18 & 19 & 17 & 17 & 18.0 & 17.8 & 19 & 0.84\\
RNN & 128 & 7 & 20 & 18 & 16 & 16 & 16 & 16 & 16.0 & 17.2 & 20 & 1.79\\
RNN & 128 & 8 & 15 & 15 & 18 & 17 & 17 & 15 & 17.0 & 16.4 & 18 & 1.34\\
\hline
\hline
\hline
GRU & 8 & 1 & 18 & 15 & 13 & 18 & 15 & 13 & 15.0 & 15.8 & 18 & 2.17\\
GRU & 8 & 2 & 22 & 20 & 21 & 17 & 18 & 17 & 20.0 & 19.6 & 22 & 2.07\\
GRU & 8 & 3 & 22 & 22 & 24 & 24 & 21 & 21 & 22.0 & 22.6 & 24 & 1.34\\
GRU & 8 & 4 & 24 & 28 & 24 & 23 & 25 & 23 & 24.0 & 24.8 & 28 & 1.92\\
GRU & 8 & 5 & 27 & 30 & 26 & 27 & 27 & 26 & 27.0 & 27.4 & 30 & 1.52\\
GRU & 8 & 6 & 26 & 26 & 23 & 22 & 23 & 22 & 23.0 & 24.0 & 26 & 1.87\\
GRU & 8 & 7 & 24 & 30 & 23 & 27 & 27 & 23 & 27.0 & 26.2 & 30 & 2.77\\
GRU & 8 & 8 & 32 & 26 & 27 & 28 & 25 & 25 & 27.0 & 27.6 & 32 & 2.70\\
\hline
GRU & 16 & 1 & 17 & 15 & 18 & 15 & 16 & 15 & 16.0 & 16.2 & 18 & 1.30\\
GRU & 16 & 2 & 20 & 17 & 17 & 20 & 17 & 17 & 17.0 & 18.2 & 20 & 1.64\\
GRU & 16 & 3 & 20 & 24 & 26 & 22 & 20 & 20 & 22.0 & 22.4 & 26 & 2.61\\
GRU & 16 & 4 & 21 & 21 & 25 & 32 & 21 & 21 & 21.0 & 24.0 & 32 & 4.80\\
GRU & 16 & 5 & 28 & 32 & 23 & 25 & 25 & 23 & 25.0 & 26.6 & 32 & 3.51\\
GRU & 16 & 6 & 32 & 29 & 27 & 34 & 26 & 26 & 29.0 & 29.6 & 34 & 3.36\\
GRU & 16 & 7 & 29 & 30 & 28 & 29 & 34 & 28 & 29.0 & 30.0 & 34 & 2.35\\
GRU & 16 & 8 & 32 & 33 & 28 & 29 & 30 & 28 & 30.0 & 30.4 & 33 & 2.07\\
    \end{tabular}
\end{table}

\clearpage

\begin{table}[]
    \centering
    \begin{tabular}{c|c|c||c|c|c|c|c||c|c|c|c|c}
        Type & Hidden Size & Layers & R1 & R2 & R3 & R4 & R5 & Min. & Med. & Mean & Max. & Std. \\
        \hline
GRU & 24 & 1 & 13 & 17 & 16 & 17 & 18 & 13 & 17.0 & 16.2 & 18 & 1.92\\
GRU & 24 & 2 & 22 & 21 & 18 & 22 & 18 & 18 & 21.0 & 20.2 & 22 & 2.05\\
GRU & 24 & 3 & 22 & 26 & 23 & 25 & 25 & 22 & 25.0 & 24.2 & 26 & 1.64\\
GRU & 24 & 4 & 27 & 22 & 25 & 22 & 26 & 22 & 25.0 & 24.4 & 27 & 2.30\\
GRU & 24 & 5 & 34 & 28 & 29 & 32 & 29 & 28 & 29.0 & 30.4 & 34 & 2.51\\
GRU & 24 & 6 & 33 & 30 & 29 & 34 & 27 & 27 & 30.0 & 30.6 & 34 & 2.88\\
GRU & 24 & 7 & 28 & 36 & 36 & 27 & 26 & 26 & 28.0 & 30.6 & 36 & 4.98\\
GRU & 24 & 8 & 27 & 30 & 32 & 32 & 27 & 27 & 30.0 & 29.6 & 32 & 2.51\\
\hline
GRU & 32 & 1 & 21 & 21 & 20 & 20 & 22 & 20 & 21.0 & 20.8 & 22 & 0.84\\
GRU & 32 & 2 & 23 & 23 & 23 & 25 & 25 & 23 & 23.0 & 23.8 & 25 & 1.10\\
GRU & 32 & 3 & 26 & 25 & 27 & 27 & 28 & 25 & 27.0 & 26.6 & 28 & 1.14\\
GRU & 32 & 4 & 29 & 32 & 29 & 32 & 31 & 29 & 31.0 & 30.6 & 32 & 1.52\\
GRU & 32 & 5 & 33 & 34 & 36 & 32 & 32 & 32 & 33.0 & 33.4 & 36 & 1.67\\
GRU & 32 & 6 & 36 & 40 & 40 & 34 & 34 & 34 & 36.0 & 36.8 & 40 & 3.03\\
GRU & 32 & 7 & 37 & 38 & 38 & 37 & 40 & 37 & 38.0 & 38.0 & 40 & 1.22\\
GRU & 32 & 8 & 37 & 38 & 37 & 37 & 39 & 37 & 37.0 & 37.6 & 39 & 0.89\\
\hline
GRU & 40 & 1 & 12 & 13 & 11 & 12 & 11 & 11 & 12.0 & 11.8 & 13 & 0.84\\
GRU & 40 & 2 & 15 & 14 & 14 & 15 & 15 & 14 & 15.0 & 14.6 & 15 & 0.55\\
GRU & 40 & 3 & 18 & 18 & 15 & 18 & 15 & 15 & 18.0 & 16.8 & 18 & 1.64\\
GRU & 40 & 4 & 20 & 18 & 17 & 20 & 20 & 17 & 20.0 & 19.0 & 20 & 1.41\\
GRU & 40 & 5 & 20 & 19 & 19 & 18 & 21 & 18 & 19.0 & 19.4 & 21 & 1.14\\
GRU & 40 & 6 & 23 & 19 & 21 & 24 & 22 & 19 & 22.0 & 21.8 & 24 & 1.92\\
GRU & 40 & 7 & 24 & 24 & 24 & 28 & 22 & 22 & 24.0 & 24.4 & 28 & 2.19\\
GRU & 40 & 8 & 26 & 23 & 27 & 26 & 25 & 23 & 26.0 & 25.4 & 27 & 1.52\\
\hline
GRU & 48 & 1 & 21 & 21 & 22 & 22 & 22 & 21 & 22.0 & 21.6 & 22 & 0.55\\
GRU & 48 & 2 & 28 & 27 & 25 & 24 & 26 & 24 & 26.0 & 26.0 & 28 & 1.58\\
GRU & 48 & 3 & 27 & 29 & 30 & 23 & 32 & 23 & 29.0 & 28.2 & 32 & 3.42\\
GRU & 48 & 4 & 27 & 35 & 40 & 35 & 31 & 27 & 35.0 & 33.6 & 40 & 4.88\\
GRU & 48 & 5 & 36 & 37 & 36 & 35 & 33 & 33 & 36.0 & 35.4 & 37 & 1.52\\
GRU & 48 & 6 & 33 & 37 & 35 & 36 & 41 & 33 & 36.0 & 36.4 & 41 & 2.97\\
GRU & 48 & 7 & 33 & 31 & 30 & 42 & 38 & 30 & 33.0 & 34.8 & 42 & 5.07\\
GRU & 48 & 8 & 35 & 38 & 36 & 38 & 39 & 35 & 38.0 & 37.2 & 39 & 1.64\\
\hline
GRU & 56 & 1 & 11 & 13 & 11 & 12 & 11 & 11 & 11.0 & 11.6 & 13 & 0.89\\
GRU & 56 & 2 & 14 & 14 & 14 & 13 & 14 & 13 & 14.0 & 13.8 & 14 & 0.45\\
GRU & 56 & 3 & 15 & 16 & 20 & 16 & 16 & 15 & 16.0 & 16.6 & 20 & 1.95\\
GRU & 56 & 4 & 21 & 20 & 21 & 21 & 19 & 19 & 21.0 & 20.4 & 21 & 0.89\\
GRU & 56 & 5 & 28 & 24 & 23 & 25 & 23 & 23 & 24.0 & 24.6 & 28 & 2.07\\
GRU & 56 & 6 & 24 & 25 & 26 & 28 & 30 & 24 & 26.0 & 26.6 & 30 & 2.41\\
GRU & 56 & 7 & 26 & 27 & 26 & 27 & 28 & 26 & 27.0 & 26.8 & 28 & 0.84\\
GRU & 56 & 8 & 29 & 27 & 28 & 30 & 27 & 27 & 28.0 & 28.2 & 30 & 1.30\\
\hline
GRU & 64 & 1 & 20 & 22 & 21 & 24 & 20 & 20 & 21.0 & 21.4 & 24 & 1.67\\
GRU & 64 & 2 & 25 & 24 & 28 & 24 & 32 & 24 & 25.0 & 26.6 & 32 & 3.44\\
GRU & 64 & 3 & 33 & 33 & 26 & 29 & 27 & 26 & 29.0 & 29.6 & 33 & 3.29\\
GRU & 64 & 4 & 32 & 36 & 31 & 23 & 31 & 23 & 31.0 & 30.6 & 36 & 4.72\\
GRU & 64 & 5 & 33 & 34 & 38 & 36 & 31 & 31 & 34.0 & 34.4 & 38 & 2.70\\
GRU & 64 & 6 & 35 & 31 & 23 & 35 & 34 & 23 & 34.0 & 31.6 & 35 & 5.08\\
GRU & 64 & 7 & 36 & 40 & 37 & 31 & 35 & 31 & 36.0 & 35.8 & 40 & 3.27\\
GRU & 64 & 8 & 38 & 38 & 40 & 40 & 42 & 38 & 40.0 & 39.6 & 42 & 1.67\\
    \end{tabular}
\end{table}

\clearpage

\begin{table}[]
    \centering
    \begin{tabular}{c|c|c||c|c|c|c|c||c|c|c|c|c}
        Type & Hidden Size & Layers & R1 & R2 & R3 & R4 & R5 & Min. & Med. & Mean & Max. & Std. \\
        \hline
GRU & 72 & 1 & 23 & 22 & 21 & 21 & 22 & 21 & 22.0 & 21.8 & 23 & 0.84\\
GRU & 72 & 2 & 23 & 29 & 23 & 24 & 21 & 21 & 23.0 & 24.0 & 29 & 3.00\\
GRU & 72 & 3 & 26 & 30 & 23 & 28 & 29 & 23 & 28.0 & 27.2 & 30 & 2.77\\
GRU & 72 & 4 & 30 & 33 & 32 & 33 & 31 & 30 & 32.0 & 31.8 & 33 & 1.30\\
GRU & 72 & 5 & 27 & 31 & 30 & 31 & 40 & 27 & 31.0 & 31.8 & 40 & 4.87\\
GRU & 72 & 6 & 31 & 31 & 34 & 33 & 34 & 31 & 33.0 & 32.6 & 34 & 1.52\\
GRU & 72 & 7 & 40 & 32 & 33 & 36 & 37 & 32 & 36.0 & 35.6 & 40 & 3.21\\
GRU & 72 & 8 & 36 & 33 & 31 & 38 & 34 & 31 & 34.0 & 34.4 & 38 & 2.70\\
\hline
GRU & 80 & 1 & 11 & 12 & 12 & 13 & 11 & 11 & 12.0 & 11.8 & 13 & 0.84\\
GRU & 80 & 2 & 15 & 13 & 16 & 15 & 13 & 13 & 15.0 & 14.4 & 16 & 1.34\\
GRU & 80 & 3 & 19 & 22 & 24 & 15 & 32 & 15 & 22.0 & 22.4 & 32 & 6.35\\
GRU & 80 & 4 & 26 & 25 & 22 & 26 & 25 & 22 & 25.0 & 24.8 & 26 & 1.64\\
GRU & 80 & 5 & 19 & 24 & 28 & 22 & 28 & 19 & 24.0 & 24.2 & 28 & 3.90\\
GRU & 80 & 6 & 29 & 27 & 27 & 27 & 28 & 27 & 27.0 & 27.6 & 29 & 0.89\\
GRU & 80 & 7 & 28 & 31 & 32 & 31 & 27 & 27 & 31.0 & 29.8 & 32 & 2.17\\
GRU & 80 & 8 & 32 & 40 & 30 & 34 & 37 & 30 & 34.0 & 34.6 & 40 & 3.97\\
\hline
GRU & 88 & 1 & 12 & 12 & 11 & 12 & 12 & 11 & 12.0 & 11.8 & 12 & 0.45\\
GRU & 88 & 2 & 15 & 13 & 13 & 15 & 13 & 13 & 13.0 & 13.8 & 15 & 1.10\\
GRU & 88 & 3 & 26 & 20 & 20 & 21 & 23 & 20 & 21.0 & 22.0 & 26 & 2.55\\
GRU & 88 & 4 & 25 & 21 & 23 & 32 & 25 & 21 & 25.0 & 25.2 & 32 & 4.15\\
GRU & 88 & 5 & 28 & 29 & 26 & 28 & 25 & 25 & 28.0 & 27.2 & 29 & 1.64\\
GRU & 88 & 6 & 32 & 29 & 32 & 25 & 28 & 25 & 29.0 & 29.2 & 32 & 2.95\\
GRU & 88 & 7 & 27 & 23 & 36 & 29 & 35 & 23 & 29.0 & 30.0 & 36 & 5.48\\
GRU & 88 & 8 & 32 & 34 & 34 & 15 & 35 & 15 & 34.0 & 30.0 & 35 & 8.46\\
\hline
GRU & 96 & 1 & 21 & 21 & 19 & 21 & 21 & 19 & 21.0 & 20.6 & 21 & 0.89\\
GRU & 96 & 2 & 23 & 25 & 22 & 23 & 19 & 19 & 23.0 & 22.4 & 25 & 2.19\\
GRU & 96 & 3 & 34 & 26 & 26 & 28 & 33 & 26 & 28.0 & 29.4 & 34 & 3.85\\
GRU & 96 & 4 & 31 & 29 & 35 & 29 & 23 & 23 & 29.0 & 29.4 & 35 & 4.34\\
GRU & 96 & 5 & 34 & 36 & 30 & 27 & 31 & 27 & 31.0 & 31.6 & 36 & 3.51\\
GRU & 96 & 6 & 36 & 31 & 33 & 33 & 27 & 27 & 33.0 & 32.0 & 36 & 3.32\\
GRU & 96 & 7 & 37 & 42 & 37 & 38 & 31 & 31 & 37.0 & 37.0 & 42 & 3.94\\
GRU & 96 & 8 & 31 & 31 & 31 & 44 & 29 & 29 & 31.0 & 33.2 & 44 & 6.10\\
\hline
GRU & 104 & 1 & 12 & 12 & 12 & 12 & 12 & 12 & 12.0 & 12.0 & 12 & 0.00\\
GRU & 104 & 2 & 15 & 15 & 14 & 16 & 14 & 14 & 15.0 & 14.8 & 16 & 0.84\\
GRU & 104 & 3 & 20 & 22 & 22 & 16 & 19 & 16 & 20.0 & 19.8 & 22 & 2.49\\
GRU & 104 & 4 & 32 & 23 & 26 & 32 & 24 & 23 & 26.0 & 27.4 & 32 & 4.34\\
GRU & 104 & 5 & 28 & 28 & 29 & 33 & 27 & 27 & 28.0 & 29.0 & 33 & 2.35\\
GRU & 104 & 6 & 30 & 28 & 40 & 38 & 35 & 28 & 35.0 & 34.2 & 40 & 5.12\\
GRU & 104 & 7 & 38 & 33 & 36 & 38 & 31 & 31 & 36.0 & 35.2 & 38 & 3.11\\
GRU & 104 & 8 & 37 & 36 & 37 & 40 & 40 & 36 & 37.0 & 38.0 & 40 & 1.87\\
\hline
GRU & 112 & 1 & 11 & 14 & 11 & 12 & 12 & 11 & 12.0 & 12.0 & 14 & 1.22\\
GRU & 112 & 2 & 14 & 14 & 14 & 16 & 14 & 14 & 14.0 & 14.4 & 16 & 0.89\\
GRU & 112 & 3 & 22 & 15 & 19 & 18 & 16 & 15 & 18.0 & 18.0 & 22 & 2.74\\
GRU & 112 & 4 & 23 & 15 & 26 & 28 & 30 & 15 & 26.0 & 24.4 & 30 & 5.86\\
GRU & 112 & 5 & 44 & 29 & 31 & 27 & 38 & 27 & 31.0 & 33.8 & 44 & 7.05\\
GRU & 112 & 6 & 33 & 46 & 30 & 34 & 37 & 30 & 34.0 & 36.0 & 46 & 6.12\\
GRU & 112 & 7 & 41 & 43 & 15 & 35 & 37 & 15 & 37.0 & 34.2 & 43 & 11.19\\
GRU & 112 & 8 & 44 & 40 & 40 & 31 & 39 & 31 & 40.0 & 38.8 & 44 & 4.76\\
    \end{tabular}
\end{table}

\begin{table}[]
    \centering
    \begin{tabular}{c|c|c||c|c|c|c|c||c|c|c|c|c}
        Type & Hidden Size & Layers & R1 & R2 & R3 & R4 & R5 & Min. & Med. & Mean & Max. & Std. \\
        \hline
GRU & 120 & 1 & 13 & 11 & 10 & 11 & 11 & 10 & 11.0 & 11.2 & 13 & 1.10\\
GRU & 120 & 2 & 13 & 13 & 13 & 14 & 14 & 13 & 13.0 & 13.4 & 14 & 0.55\\
GRU & 120 & 3 & 22 & 15 & 21 & 24 & 15 & 15 & 21.0 & 19.4 & 24 & 4.16\\
GRU & 120 & 4 & 28 & 15 & 15 & 30 & 27 & 15 & 27.0 & 23.0 & 30 & 7.38\\
GRU & 120 & 5 & 28 & 31 & 15 & 40 & 31 & 15 & 31.0 & 29.0 & 40 & 9.03\\
GRU & 120 & 6 & 38 & 35 & 30 & 35 & 50 & 30 & 35.0 & 37.6 & 50 & 7.50\\
GRU & 120 & 7 & 33 & 48 & 31 & 40 & 40 & 31 & 40.0 & 38.4 & 48 & 6.73\\
GRU & 120 & 8 & 40 & 40 & 44 & 44 & 44 & 40 & 44.0 & 42.4 & 44 & 2.19\\
\hline
GRU & 128 & 1 & 20 & 24 & 20 & 18 & 20 & 18 & 20.0 & 20.4 & 24 & 2.19\\
GRU & 128 & 2 & 27 & 22 & 23 & 24 & 19 & 19 & 23.0 & 23.0 & 27 & 2.92\\
GRU & 128 & 3 & 22 & 27 & 30 & 25 & 32 & 22 & 27.0 & 27.2 & 32 & 3.96\\
GRU & 128 & 4 & 23 & 30 & 27 & 27 & 29 & 23 & 27.0 & 27.2 & 30 & 2.68\\
GRU & 128 & 5 & 33 & 42 & 33 & 48 & 40 & 33 & 40.0 & 39.2 & 48 & 6.38\\
GRU & 128 & 6 & 29 & 29 & 30 & 31 & 33 & 29 & 30.0 & 30.4 & 33 & 1.67\\
GRU & 128 & 7 & 48 & 27 & 30 & 37 & 38 & 27 & 37.0 & 36.0 & 48 & 8.15\\
GRU & 128 & 8 & 33 & 31 & 36 & 41 & 40 & 31 & 36.0 & 36.2 & 41 & 4.32\\
\hline
\hline
\hline
LSTM & 8 & 1 & 16 & 16 & 16 & 14 & 13 & 13 & 16.0 & 15.0 & 16 & 1.41\\
LSTM & 8 & 2 & 15 & 19 & 18 & 16 & 16 & 15 & 16.0 & 16.8 & 19 & 1.64\\
LSTM & 8 & 3 & 18 & 24 & 20 & 18 & 18 & 18 & 18.0 & 19.6 & 24 & 2.61\\
LSTM & 8 & 4 & 20 & 17 & 15 & 17 & 15 & 15 & 17.0 & 16.8 & 20 & 2.05\\
LSTM & 8 & 5 & 22 & 15 & 20 & 16 & 16 & 15 & 16.0 & 17.8 & 22 & 3.03\\
LSTM & 8 & 6 & 14 & 17 & 14 & 15 & 17 & 14 & 15.0 & 15.4 & 17 & 1.52\\
LSTM & 8 & 7 & 11 & 15 & 13 & 13 & 13 & 11 & 13.0 & 13.0 & 15 & 1.41\\
LSTM & 8 & 8 & 0 & 0 & 0 & 0 & 0 & 0 & 0.0 & 0.0 & 0 & 0.00\\
\hline
LSTM & 16 & 1 & 14 & 14 & 14 & 16 & 16 & 14 & 14.0 & 14.8 & 16 & 1.10\\
LSTM & 16 & 2 & 15 & 16 & 22 & 15 & 15 & 15 & 15.0 & 16.6 & 22 & 3.05\\
LSTM & 16 & 3 & 18 & 23 & 22 & 21 & 32 & 18 & 22.0 & 23.2 & 32 & 5.26\\
LSTM & 16 & 4 & 21 & 19 & 20 & 22 & 15 & 15 & 20.0 & 19.4 & 22 & 2.70\\
LSTM & 16 & 5 & 22 & 20 & 24 & 17 & 20 & 17 & 20.0 & 20.6 & 24 & 2.61\\
LSTM & 16 & 6 & 18 & 19 & 17 & 15 & 18 & 15 & 18.0 & 17.4 & 19 & 1.52\\
LSTM & 16 & 7 & 19 & 18 & 18 & 17 & 20 & 17 & 18.0 & 18.4 & 20 & 1.14\\
LSTM & 16 & 8 & 19 & 19 & 17 & 17 & 17 & 17 & 17.0 & 17.8 & 19 & 1.10\\
\hline
LSTM & 24 & 1 & 13 & 14 & 18 & 16 & 13 & 13 & 14.0 & 14.8 & 18 & 2.17\\
LSTM & 24 & 2 & 20 & 20 & 20 & 20 & 21 & 20 & 20.0 & 20.2 & 21 & 0.45\\
LSTM & 24 & 3 & 19 & 22 & 30 & 20 & 20 & 19 & 20.0 & 22.2 & 30 & 4.49\\
LSTM & 24 & 4 & 19 & 19 & 23 & 21 & 23 & 19 & 21.0 & 21.0 & 23 & 2.00\\
LSTM & 24 & 5 & 21 & 32 & 24 & 22 & 22 & 21 & 22.0 & 24.2 & 32 & 4.49\\
LSTM & 24 & 6 & 18 & 22 & 20 & 21 & 19 & 18 & 20.0 & 20.0 & 22 & 1.58\\
LSTM & 24 & 7 & 20 & 17 & 24 & 18 & 20 & 17 & 20.0 & 19.8 & 24 & 2.68\\
LSTM & 24 & 8 & 24 & 22 & 19 & 19 & 24 & 19 & 22.0 & 21.6 & 24 & 2.51\\
\hline
LSTM & 32 & 1 & 18 & 15 & 15 & 20 & 19 & 15 & 18.0 & 17.4 & 20 & 2.30\\
LSTM & 32 & 2 & 25 & 28 & 23 & 18 & 25 & 18 & 25.0 & 23.8 & 28 & 3.70\\
LSTM & 32 & 3 & 25 & 26 & 32 & 32 & 23 & 23 & 26.0 & 27.6 & 32 & 4.16\\
LSTM & 32 & 4 & 22 & 26 & 25 & 19 & 23 & 19 & 23.0 & 23.0 & 26 & 2.74\\
LSTM & 32 & 5 & 23 & 25 & 21 & 33 & 26 & 21 & 25.0 & 25.6 & 33 & 4.56\\
LSTM & 32 & 6 & 25 & 19 & 23 & 22 & 25 & 19 & 23.0 & 22.8 & 25 & 2.49\\
LSTM & 32 & 7 & 24 & 18 & 19 & 23 & 24 & 18 & 23.0 & 21.6 & 24 & 2.88\\
LSTM & 32 & 8 & 22 & 24 & 20 & 29 & 24 & 20 & 24.0 & 23.8 & 29 & 3.35\\
    \end{tabular}
\end{table}

\begin{table}[]
    \centering
    \begin{tabular}{c|c|c||c|c|c|c|c||c|c|c|c|c}
        Type & Hidden Size & Layers & R1 & R2 & R3 & R4 & R5 & Min. & Med. & Mean & Max. & Std. \\
        \hline
LSTM & 40 & 1 & 8 & 7 & 7 & 7 & 6 & 6 & 7.0 & 7.0 & 8 & 0.71\\
LSTM & 40 & 2 & 9 & 8 & 8 & 10 & 8 & 8 & 8.0 & 8.6 & 10 & 0.89\\
LSTM & 40 & 3 & 9 & 11 & 11 & 11 & 11 & 9 & 11.0 & 10.6 & 11 & 0.89\\
LSTM & 40 & 4 & 14 & 12 & 13 & 12 & 13 & 12 & 13.0 & 12.8 & 14 & 0.84\\
LSTM & 40 & 5 & 14 & 15 & 16 & 15 & 15 & 14 & 15.0 & 15.0 & 16 & 0.71\\
LSTM & 40 & 6 & 7 & 8 & 7 & 7 & 7 & 7 & 7.0 & 7.2 & 8 & 0.45\\
LSTM & 40 & 7 & 7 & 8 & 9 & 9 & 8 & 7 & 8.0 & 8.2 & 9 & 0.84\\
LSTM & 40 & 8 & 9 & 7 & 7 & 9 & 8 & 7 & 8.0 & 8.0 & 9 & 1.00\\
\hline
LSTM & 48 & 1 & 9 & 9 & 9 & 8 & 9 & 8 & 9.0 & 8.8 & 9 & 0.45\\
LSTM & 48 & 2 & 11 & 11 & 11 & 12 & 10 & 10 & 11.0 & 11.0 & 12 & 0.71\\
LSTM & 48 & 3 & 15 & 15 & 13 & 13 & 14 & 13 & 14.0 & 14.0 & 15 & 1.00\\
LSTM & 48 & 4 & 17 & 17 & 17 & 16 & 16 & 16 & 17.0 & 16.6 & 17 & 0.55\\
LSTM & 48 & 5 & 20 & 17 & 19 & 20 & 18 & 17 & 19.0 & 18.8 & 20 & 1.30\\
LSTM & 48 & 6 & 20 & 18 & 21 & 20 & 20 & 18 & 20.0 & 19.8 & 21 & 1.10\\
LSTM & 48 & 7 & 11 & 15 & 21 & 22 & 22 & 11 & 21.0 & 18.2 & 22 & 4.97\\
LSTM & 48 & 8 & 11 & 11 & 14 & 14 & 14 & 11 & 14.0 & 12.8 & 14 & 1.64\\
\hline
LSTM & 56 & 1 & 8 & 9 & 9 & 9 & 9 & 8 & 9.0 & 8.8 & 9 & 0.45\\
LSTM & 56 & 2 & 11 & 12 & 12 & 12 & 11 & 11 & 12.0 & 11.6 & 12 & 0.55\\
LSTM & 56 & 3 & 14 & 14 & 14 & 16 & 14 & 14 & 14.0 & 14.4 & 16 & 0.89\\
LSTM & 56 & 4 & 17 & 15 & 19 & 17 & 20 & 15 & 17.0 & 17.6 & 20 & 1.95\\
LSTM & 56 & 5 & 19 & 22 & 19 & 20 & 19 & 19 & 19.0 & 19.8 & 22 & 1.30\\
LSTM & 56 & 6 & 22 & 22 & 15 & 22 & 19 & 15 & 22.0 & 20.0 & 22 & 3.08\\
LSTM & 56 & 7 & 12 & 11 & 16 & 16 & 12 & 11 & 12.0 & 13.4 & 16 & 2.41\\
LSTM & 56 & 8 & 12 & 14 & 10 & 12 & 12 & 10 & 12.0 & 12.0 & 14 & 1.41\\
\hline
LSTM & 64 & 1 & 19 & 17 & 17 & 19 & 19 & 17 & 19.0 & 18.2 & 19 & 1.10\\
LSTM & 64 & 2 & 25 & 32 & 20 & 20 & 22 & 20 & 22.0 & 23.8 & 32 & 5.02\\
LSTM & 64 & 3 & 32 & 24 & 20 & 20 & 24 & 20 & 24.0 & 24.0 & 32 & 4.90\\
LSTM & 64 & 4 & 23 & 28 & 28 & 28 & 30 & 23 & 28.0 & 27.4 & 30 & 2.61\\
LSTM & 64 & 5 & 26 & 28 & 26 & 25 & 23 & 23 & 26.0 & 25.6 & 28 & 1.82\\
LSTM & 64 & 6 & 28 & 24 & 28 & 23 & 25 & 23 & 25.0 & 25.6 & 28 & 2.30\\
LSTM & 64 & 7 & 25 & 19 & 26 & 32 & 36 & 19 & 26.0 & 27.6 & 36 & 6.58\\
LSTM & 64 & 8 & 30 & 26 & 23 & 31 & 23 & 23 & 26.0 & 26.6 & 31 & 3.78\\
\hline
LSTM & 72 & 1 & 12 & 11 & 12 & 12 & 12 & 11 & 12.0 & 11.8 & 12 & 0.45\\
LSTM & 72 & 2 & 15 & 15 & 15 & 15 & 16 & 15 & 15.0 & 15.2 & 16 & 0.45\\
LSTM & 72 & 3 & 15 & 15 & 17 & 17 & 16 & 15 & 16.0 & 16.0 & 17 & 1.00\\
LSTM & 72 & 4 & 17 & 18 & 17 & 17 & 16 & 16 & 17.0 & 17.0 & 18 & 0.71\\
LSTM & 72 & 5 & 22 & 20 & 20 & 19 & 20 & 19 & 20.0 & 20.2 & 22 & 1.10\\
LSTM & 72 & 6 & 23 & 22 & 23 & 21 & 22 & 21 & 22.0 & 22.2 & 23 & 0.84\\
LSTM & 72 & 7 & 24 & 23 & 22 & 23 & 24 & 22 & 23.0 & 23.2 & 24 & 0.84\\
LSTM & 72 & 8 & 26 & 15 & 24 & 24 & 25 & 15 & 24.0 & 22.8 & 26 & 4.44\\
\hline
LSTM & 80 & 1 & 10 & 12 & 11 & 10 & 10 & 10 & 10.0 & 10.6 & 12 & 0.89\\
LSTM & 80 & 2 & 13 & 15 & 16 & 14 & 14 & 13 & 14.0 & 14.4 & 16 & 1.14\\
LSTM & 80 & 3 & 19 & 17 & 16 & 16 & 20 & 16 & 17.0 & 17.6 & 20 & 1.82\\
LSTM & 80 & 4 & 18 & 20 & 18 & 19 & 20 & 18 & 19.0 & 19.0 & 20 & 1.00\\
LSTM & 80 & 5 & 17 & 19 & 19 & 22 & 21 & 17 & 19.0 & 19.6 & 22 & 1.95\\
LSTM & 80 & 6 & 24 & 22 & 19 & 24 & 20 & 19 & 22.0 & 21.8 & 24 & 2.28\\
LSTM & 80 & 7 & 15 & 15 & 22 & 20 & 14 & 14 & 15.0 & 17.2 & 22 & 3.56\\
LSTM & 80 & 8 & 17 & 17 & 15 & 14 & 15 & 14 & 15.0 & 15.6 & 17 & 1.34\\
    \end{tabular}
\end{table}

\begin{table}[]
    \centering
    \begin{tabular}{c|c|c||c|c|c|c|c||c|c|c|c|c}
        Type & Hidden Size & Layers & R1 & R2 & R3 & R4 & R5 & Min. & Med. & Mean & Max. & Std. \\
        \hline
LSTM & 88 & 1 & 11 & 13 & 10 & 12 & 11 & 10 & 11.0 & 11.4 & 13 & 1.14\\
LSTM & 88 & 2 & 15 & 16 & 15 & 14 & 16 & 14 & 15.0 & 15.2 & 16 & 0.84\\
LSTM & 88 & 3 & 17 & 17 & 18 & 17 & 17 & 17 & 17.0 & 17.2 & 18 & 0.45\\
LSTM & 88 & 4 & 20 & 19 & 19 & 20 & 20 & 19 & 20.0 & 19.6 & 20 & 0.55\\
LSTM & 88 & 5 & 22 & 18 & 22 & 19 & 22 & 18 & 22.0 & 20.6 & 22 & 1.95\\
LSTM & 88 & 6 & 22 & 21 & 22 & 20 & 19 & 19 & 21.0 & 20.8 & 22 & 1.30\\
LSTM & 88 & 7 & 20 & 24 & 15 & 18 & 20 & 15 & 20.0 & 19.4 & 24 & 3.29\\
LSTM & 88 & 8 & 16 & 14 & 20 & 11 & 15 & 11 & 15.0 & 15.2 & 20 & 3.27\\
\hline
LSTM & 96 & 1 & 17 & 24 & 18 & 17 & 22 & 17 & 18.0 & 19.6 & 24 & 3.21\\
LSTM & 96 & 2 & 24 & 20 & 20 & 25 & 19 & 19 & 20.0 & 21.6 & 25 & 2.70\\
LSTM & 96 & 3 & 28 & 23 & 26 & 17 & 27 & 17 & 26.0 & 24.2 & 28 & 4.44\\
LSTM & 96 & 4 & 26 & 28 & 26 & 32 & 30 & 26 & 28.0 & 28.4 & 32 & 2.61\\
LSTM & 96 & 5 & 23 & 23 & 26 & 25 & 24 & 23 & 24.0 & 24.2 & 26 & 1.30\\
LSTM & 96 & 6 & 23 & 29 & 28 & 23 & 34 & 23 & 28.0 & 27.4 & 34 & 4.62\\
LSTM & 96 & 7 & 30 & 25 & 29 & 26 & 31 & 25 & 29.0 & 28.2 & 31 & 2.59\\
LSTM & 96 & 8 & 29 & 23 & 33 & 36 & 27 & 23 & 29.0 & 29.6 & 36 & 5.08\\
\hline
LSTM & 104 & 1 & 12 & 11 & 11 & 12 & 14 & 11 & 12.0 & 12.0 & 14 & 1.22\\
LSTM & 104 & 2 & 16 & 16 & 16 & 15 & 15 & 15 & 16.0 & 15.6 & 16 & 0.55\\
LSTM & 104 & 3 & 19 & 20 & 18 & 19 & 18 & 18 & 19.0 & 18.8 & 20 & 0.84\\
LSTM & 104 & 4 & 18 & 20 & 21 & 20 & 17 & 17 & 20.0 & 19.2 & 21 & 1.64\\
LSTM & 104 & 5 & 19 & 20 & 21 & 21 & 18 & 18 & 20.0 & 19.8 & 21 & 1.30\\
LSTM & 104 & 6 & 20 & 20 & 23 & 24 & 22 & 20 & 22.0 & 21.8 & 24 & 1.79\\
LSTM & 104 & 7 & 20 & 18 & 21 & 16 & 15 & 15 & 18.0 & 18.0 & 21 & 2.55\\
LSTM & 104 & 8 & 16 & 15 & 11 & 18 & 24 & 11 & 16.0 & 16.8 & 24 & 4.76\\
\hline
LSTM & 112 & 1 & 12 & 13 & 11 & 12 & 11 & 11 & 12.0 & 11.8 & 13 & 0.84\\
LSTM & 112 & 2 & 15 & 17 & 17 & 15 & 16 & 15 & 16.0 & 16.0 & 17 & 1.00\\
LSTM & 112 & 3 & 15 & 15 & 17 & 17 & 16 & 15 & 16.0 & 16.0 & 17 & 1.00\\
LSTM & 112 & 4 & 21 & 22 & 20 & 20 & 19 & 19 & 20.0 & 20.4 & 22 & 1.14\\
LSTM & 112 & 5 & 21 & 23 & 23 & 20 & 21 & 20 & 21.0 & 21.6 & 23 & 1.34\\
LSTM & 112 & 6 & 20 & 23 & 22 & 22 & 26 & 20 & 22.0 & 22.6 & 26 & 2.19\\
LSTM & 112 & 7 & 28 & 19 & 26 & 25 & 23 & 19 & 25.0 & 24.2 & 28 & 3.42\\
LSTM & 112 & 8 & 7 & 15 & 17 & 18 & 7 & 7 & 15.0 & 12.8 & 18 & 5.40\\
\hline
LSTM & 120 & 1 & 11 & 12 & 12 & 12 & 12 & 11 & 12.0 & 11.8 & 12 & 0.45\\
LSTM & 120 & 2 & 16 & 18 & 17 & 15 & 16 & 15 & 16.0 & 16.4 & 18 & 1.14\\
LSTM & 120 & 3 & 17 & 19 & 19 & 18 & 18 & 17 & 18.0 & 18.2 & 19 & 0.84\\
LSTM & 120 & 4 & 19 & 21 & 20 & 19 & 19 & 19 & 19.0 & 19.6 & 21 & 0.89\\
LSTM & 120 & 5 & 21 & 21 & 23 & 21 & 19 & 19 & 21.0 & 21.0 & 23 & 1.41\\
LSTM & 120 & 6 & 26 & 23 & 24 & 22 & 26 & 22 & 24.0 & 24.2 & 26 & 1.79\\
LSTM & 120 & 7 & 22 & 24 & 23 & 19 & 26 & 19 & 23.0 & 22.8 & 26 & 2.59\\
LSTM & 120 & 8 & 24 & 19 & 16 & 15 & 18 & 15 & 18.0 & 18.4 & 24 & 3.51\\
\hline
LSTM & 128 & 1 & 18 & 20 & 20 & 19 & 20 & 18 & 20.0 & 19.4 & 20 & 0.89\\
LSTM & 128 & 2 & 25 & 21 & 19 & 24 & 22 & 19 & 22.0 & 22.2 & 25 & 2.39\\
LSTM & 128 & 3 & 22 & 23 & 19 & 23 & 23 & 19 & 23.0 & 22.0 & 23 & 1.73\\
LSTM & 128 & 4 & 26 & 21 & 26 & 24 & 24 & 21 & 24.0 & 24.2 & 26 & 2.05\\
LSTM & 128 & 5 & 21 & 26 & 32 & 28 & 28 & 21 & 28.0 & 27.0 & 32 & 4.00\\
LSTM & 128 & 6 & 23 & 32 & 29 & 27 & 32 & 23 & 29.0 & 28.6 & 32 & 3.78\\
LSTM & 128 & 7 & 23 & 32 & 28 & 27 & 27 & 23 & 27.0 & 27.4 & 32 & 3.21\\
LSTM & 128 & 8 & 30 & 32 & 25 & 34 & 32 & 25 & 32.0 & 30.6 & 34 & 3.44\\
    \end{tabular}
\end{table}

\end{document}